\documentclass[lettersize,journal,twoside, 10pt]{IEEEtran}  
\usepackage{cite}
\usepackage{amsmath,amssymb,amsfonts}
\usepackage{algorithmic}
\usepackage{graphicx}
\usepackage{textcomp}
\usepackage{wrapfig}
\usepackage{hyperref}

\usepackage{cite}
\usepackage{caption}
\usepackage{multirow}
\usepackage{lipsum}
\usepackage{booktabs}
\usepackage{tabularx}
\newcolumntype{P}[1]{>{\centering\arraybackslash}p{#1}}
\newcolumntype{M}[1]{>{\centering\arraybackslash}m{#1}}
\newcolumntype{C}[1]{>{\centering\arraybackslash}m{#1}}

\begin{document}
\title{A Machine Learning Approach to Sensor Substitution from Tactile Sensing to Visual Perception for Non-Prehensile Manipulation} 
\author{Idil Ozdamar, Doganay Sirintuna, and Arash Ajoudani
\thanks{Received 8 May 2025. This work was supported in part by the European Union Horizon Project TORNADO (GA 101189557). \textit{(Idil Ozdamar and Doganay Sirintuna contributed equally to this work.) (Corresponding author: Idil Ozdamar.)}% This paragraph of the first footnote will contain the date on 
% which you submitted your paper for review. It will also contain support 
% information, including sponsor and financial support acknowledgment. For 
% example, ``This work was supported in part by the U.S. Department of 
% Commerce under Grant BS123456.'' }
}
\thanks{The authors are with the HRI$^2$ Lab, Istituto Italiano di Tecnologia, Genoa, Italy. Idil Ozdamar and Doganay Sirintuna are also with Dept. of Informatics, Bioengineering, Robotics, and System Engineering. University of Genoa, Genoa, Italy (e-mails: idil.ozdamar@iit.it; doganay.sirintuna@iit.it;arash.ajoudani@iit.it).}}

\maketitle

\begin{abstract}
Mobile manipulators are increasingly deployed in complex environments, requiring diverse sensors to perceive and interact with their surroundings. However, equipping every robot with every possible sensor is often impractical due to cost and physical constraints.  A critical challenge arises when robots with differing sensor capabilities need to collaborate or perform similar tasks.  For example, consider a scenario where a mobile manipulator equipped with high-resolution tactile skin is skilled at non-prehensile manipulation tasks like pushing.  If this robot needs to be replaced or augmented by a robot lacking such tactile sensing, the learned manipulation policies become inapplicable. This paper addresses the problem of sensor substitution in non-prehensile manipulation. We propose a novel machine learning-based framework that enables a robot with a limited sensor set (e.g., LiDAR or RGB-D) to effectively perform tasks previously reliant on a richer sensor suite (e.g., tactile skin). Our approach learns a mapping between the available sensor data and the information provided by the substituted sensor, effectively synthesizing the missing sensory input. Specifically, we demonstrate the efficacy of our framework by training a model to substitute tactile skin data for the task of non-prehensile pushing using a mobile manipulator.  We show that a manipulator equipped only with LiDAR or RGB-D can, after training, achieve comparable and sometimes even better pushing performance to a mobile base utilizing direct tactile feedback.  
\end{abstract}

\begin{IEEEkeywords}
Cross-modal sensor substitution via deep learning, Long Short-Term Memory (LSTM), reactive non-prehensile manipulation, sim-to-real transfer, tactile sensing, visual perception.
\end{IEEEkeywords}

\section{Introduction}
\label{sec:introduction}
\IEEEPARstart{M}{obile} manipulators are increasingly deployed in complex environments, requiring diverse sensor suites for effective perception and interaction.  However, practical constraints like cost, space, and power consumption often limit the sensors a robot can carry. This sensor heterogeneity creates a significant challenge, especially when robots need to collaborate or perform similar tasks.  Specifically, if a robot relies on a particular sensory modality for a task, and that modality is unavailable on a collaborating or replacement robot, the task may become impossible.

\begin{figure}[!t]	
        \centering
	\includegraphics[width=0.99\linewidth]{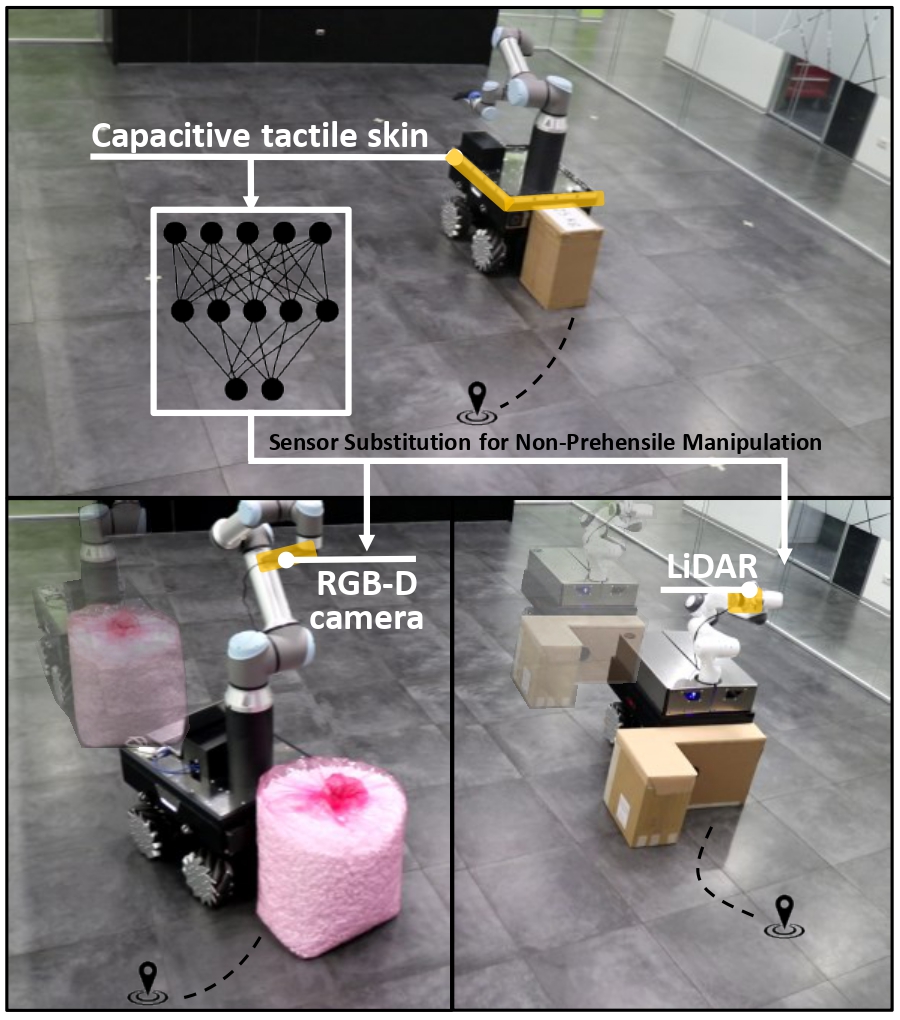}
	\caption{This study introduces a novel sensor substitution approach from tactile sensing to visual perception that enables non-prehensile manipulation of objects with varying physical characteristics, even when the original sensory information (e.g., tactile sensor) used by the mobile manipulators is unavailable.}
	\label{fig:digest}
\end{figure}

Cross-modal sensory transformation offers a promising solution to this problem by enabling the use of alternative sensory information when the primary modality is absent. This process involves acquiring data from available sensors, extracting relevant features, and transforming them into a format compatible with the target perception system.  A foundational example of this concept is the work presented in \cite{RitaSensorySubsDef}, which pioneered the use of tactile stimuli to infer visual information for visually impaired individuals. This work laid the groundwork for sensor substitution research, including its application to robotic systems. By enabling robots to exploit alternative sensory inputs while potentially maintaining the same control strategies, cross-modal transformation can significantly enhance their versatility and adaptability.

In robotic manipulation, the environment is primarily perceived through sensory modalities based on point clouds, vision, or touch, which complement each other \cite{surveyMultimodal}. 
Vision and point cloud-based perceptions provide rich spatial and structural information, while touch offers detailed feedback about contact dynamics.
This complementarity has made their fusion a popular research topic, as it allows for transforming raw data into forms that can be meaningfully combined \cite{multiBjörkman,multiIlonen,multiLiu,multiModKroemervisionTouch}. 
However, this approach is not suitable for replacing one sensory modality with another, leaving the challenge of cross-modal transformation — where one modality compensates for the absence of the other — largely unexplored.

Li et al. \cite{LiYunzhu} highlighted this gap in their study as well, noting that the cross-modal transformation between vision and touch remained unaddressed due to the lack of sufficient datasets. To overcome this limitation, they automated data collection and showed that conditional adversarial networks can predict 2D vision from touch and vice versa. 
Instead of using 2D visual input, Falco et al. \cite{Falco} introduced a transfer learning approach that uses point cloud-based object representations, incorporating feature extraction and domain adaptation to enable vision-to-tactile object recognition through shape.
Building on this, Murali et al. \cite{Murali} advanced visuotactile cross-modal object recognition through deep active transfer learning, training the network on dense point clouds and enabling recognition with sparse tactile point clouds.
Collectively, these works demonstrated the feasibility of cross-modal mapping using machine learning techniques, although it has not yet been directly extended to practical robotic applications.

Aligned with the goal of advancing manipulation capabilities on the robots, non-prehensile manipulation stands out as a fundamental motion primitive \cite{survey}. It becomes particularly vital, especially for tasks like clearing paths \cite{taskin} or transporting unwieldy and ungraspable objects \cite{unwieldy}. While non-prehensile manipulation, such as pushing with body parts like legs, arms, etc., is intuitive for humans, designing controllers for robots is considerably challenging due to the diverse and non-uniform properties of objects \cite{LetsPushReview}.

To address this challenge, researchers have developed various strategies, often relying on frictional simplifications and a priori knowledge about the object, which are difficult to exploit in real-world pushing scenarios. For instance, Lynch and Mason \cite{Lynch_Mason} presented an open-loop planner for stable pushing by enforcing sticking line contact, assuming a known friction value and uniform friction properties over the support plane. This idea was extended in \cite{Zhou2019PushingRevisited} for single-point contact scenarios under similar assumptions. Since the sticking constraint enforces a fixed relative position between the robot and the object, stable pushing eliminates repositioning actions, which are hard to execute with nonholonomic robots. Similarly, Bertoncelli et al. \cite{Bertoncelli2020} incorporated these stability constraints into an MPC framework, and later,  \cite{unwieldy} simplified this controller for computational efficiency.

Contrary to the studies mentioned above that presume object-specific properties, another line of research is dedicated to pushing the objects using reactive motions based on sensory feedback. Lau et al. \cite{Lau2011} learned object behavior from the marker-based vision system and then utilized it to push the irregular-shaped objects to the desired direction. Another approach \cite{Igarashi2010} relied on marker-based tracking to propose an adaptive controller that combines orbiting around the object and pushing motion for a mobile robot with a circular base. This concept was improved and validated in \cite{Pushing_corridor}, where the inverse model of the unknown object interaction is used in real-time for improved manipulation. On the other hand, Lloyd and Lepora \cite{LoydAndLepora} demonstrated that an optical tactile sensor attached to the end-effector, providing local object information, can also be employed to push objects toward targets, offering an alternative to vision-based measurements.

Building on this line of research, the prior work \cite{idilPushDark} introduced a non-prehensile manipulation strategy that generates reactive pushing maneuvers based on the contact location information obtained from a novel tactile sensor that covers the sides of the mobile base. This approach outperforms the strategy presented in~\cite{Pushing_corridor}, the closest state-of-the-art mobile pushing technique comparable to the one described here, as both methods do not depend on predefined assumptions about object-specific characteristics or pre-modeled object behaviors. Although the strategies presented so far enable push manipulation by leveraging vision and tactile sensory feedback, their effectiveness remained limited by the hardware upon which the controller strategy was based. At this point, employing a sensor substitution approach without compromising controller performance would not only improve adaptability to diverse robotic platforms but also significantly broaden potential deployment scenarios with varying sensor availability.

In this study, we introduce a novel sensor substitution approach using machine learning techniques to enable non-prehensile manipulation in the absence of the original sensory information on which the pushing strategy was built (see Fig. \ref{fig:digest}). We have demonstrated that the \emph{Reactive Pushing Strategy} (\textit{RPS}) \cite{idilPushDark}, which relied on the tactile feedback, can be successful enough to push objects with varying shapes, frictional properties, and inertial characteristics using different sensor modalities. The key contributions of this study can be outlined as follows:
\begin{itemize}
    \item A systematic cross-modal transformation approach is proposed by leveraging Long Short-Term Memory networks to deduce contact information. Central to this approach is an innovative descriptor that maintains informational richness while enabling real-time operation.
    \item A data collection and training pipeline for the \emph{Contact Perception Module (CPM)} is introduced in a simulation environment by using the data obtained from both the 3D LiDAR sensor attached at the end-effector and the tactile sensor covering the base while the robot is pushing different objects with varying shapes, frictions, and inertial properties to target points based on the contact location.
    \item A systematic experimental evaluation of the \emph{CPM} in a simulation environment is presented, where the \textit{RPS} attempts to push objects solely using the 3D LiDAR sensor to reach target points, both those included in the training and previously unseen ones. Furthermore, the scalability of the proposed approach was also validated by using an RGB-D camera without the need for retraining, thereby demonstrating the system's ability to handle hardware variability.
    \item A comprehensive real-world experimentation was conducted, demonstrating that our approach provides a practical solution for enabling the use of an alternative sensor when the original is missing. In \cite{idilPushDark}, the inherent requirement of the capacitive-based tactile sensor to surpass a certain pressure threshold limited its reliability in capturing contact interactions with lightweight or deformable objects, leaving the challenge of manipulating such objects unresolved. In contrast, the proposed sensor substitution approach successfully pushed a broader range of objects, spanning from rigid to deformable, light to heavy, and convex to concave, indicating a significant improvement in adaptability without requiring sim-to-real tuning.
\end{itemize}

\begin{figure*}[!t]	
        \centering
	\includegraphics[width=0.99\textwidth]{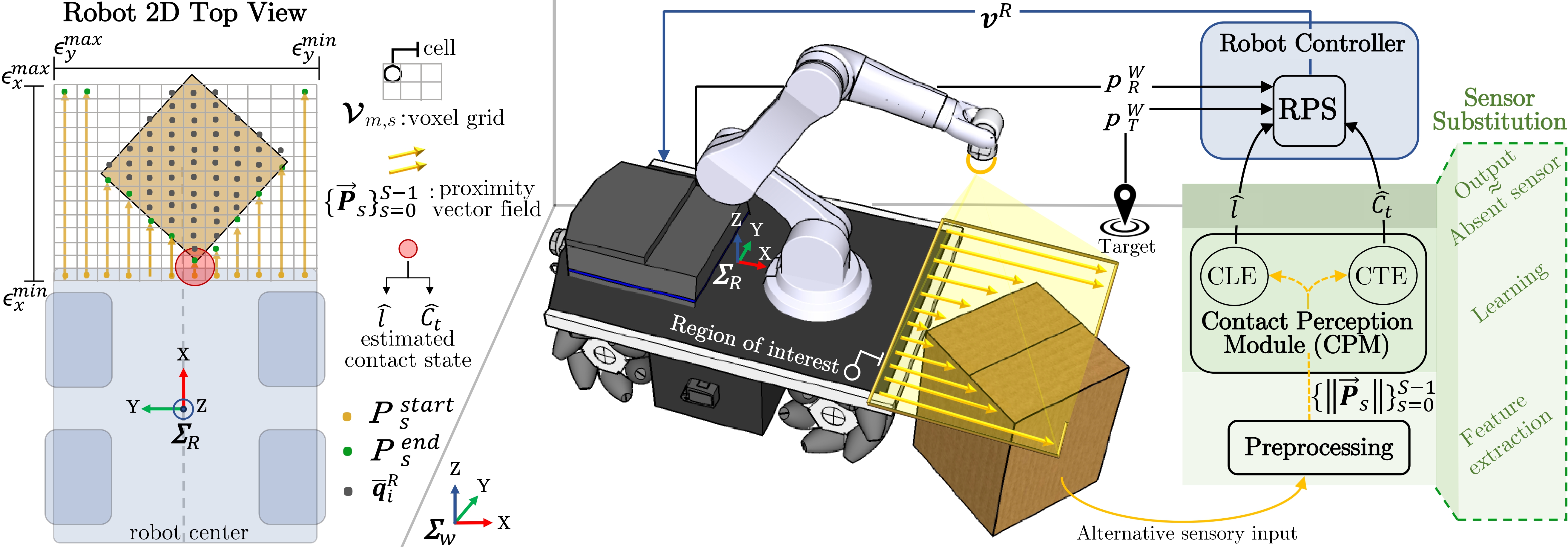}
	\caption{High-level scheme of the proposed framework.}
	\label{fig:framework}
\end{figure*}

\section{Proposed Framework}
\subsection{Robotic Platforms and Perception Modalities}\label{sec:robot_platform}

In this study, we utilized two different perception modalities for substitution in mobile manipulators. The original study \cite{idilPushDark} utilized a capacitive touch sensor covering the mobile base, which has now been replaced by a 3D LiDAR (Ouster OS0-32 with a $\pm$ 45° field of view) mounted on the end-effector to capture the point cloud, ensuring coverage of the area where the base performs non-prehensile manipulation. To demonstrate scalability across sensing modalities, we also exploit an RGB-D camera (Intel RealSense D435i) mounted on the end-effector (see Sec. \ref{sec:framework_scalability}). The mobile manipulators used for the experiments have the same omnidirectional active base (Robotnik SUMMIT-XL STEEL) with a 250 kg payload capacity that eliminates non-holonomic constraints, enabling unrestricted movement across an expansive workspace while being well-suited for pushing large, bulky objects.  Since the robotic arms remain static at a predefined position for sensor placement and are not actively involved in pushing, their specifications are not provided.

\subsection{Reactive Pushing Strategy (RPS)}\label{sec:reactive_pushing_strategy}

In \cite{idilPushDark}, a novel pushing strategy is introduced (\textit{RPS}) for mobile robots to drive objects toward a target location by utilizing the contact location alone, without relying on additional information sources. This strategy aims to maintain the object close to the center of the robot by dynamically adjusting the motion while progressing toward the target. According to this, the desired linear velocities in X and Y directions, denoted as ${v}^{R}_x$ and ${v}^{R}_y$ $\in$ $\mathbb{R}$, are generated using the parameter, $v^{*}$ $\in$ $\mathbb{R}_{\geq 0 }$, which is determined by the remaining distance to target, and the adaptive rate, $a_{r}$ $\in$ $\mathbb{R}$, which is derived from the contact location. Firstly, the $v^{*}$ is formulated as follows:
\begin{equation}\label{eq:v_star}
v^{*} = K_{v}||\boldsymbol{d}||, \quad \boldsymbol{d}  = \boldsymbol{p}^{W}_{T} - (\boldsymbol{p}^{W}_{R} + \boldsymbol{R}^{W}_{R}\boldsymbol{p}^{R}_{C})
\end{equation}
where $\boldsymbol{d}$ $\in$ $\mathbb{R}^{2}$ is the displacement vector, $K_{v}$ $\in$ $\mathbb{R}_{>0}$ is the velocity gain, $\boldsymbol{p}^{W}_{T}$ and $\boldsymbol{p}^{W}_{R}$ $\in$ $\mathbb{R}^{2}$ represent the target and robot positions w.r.t the world frame, $\boldsymbol{R}^{W}_{R}$ $\in$ $\mathbb{R}^{2\times 2}$ is the rotation matrix between the world ($\boldsymbol{\Sigma}_{W}$) and the robot base frame ($\boldsymbol{\Sigma}_{R}$). Secondly, the adaptive rate, $a_{r}$, is calculated based on the current contact location during the pushing process using a logistic function, as follows: 
\begin{equation} \label{eq:a_r}
    a_{r} = \begin{cases} (\zeta + \frac{(\eta - \zeta)}{1+e^{(\beta-|{l}|)k}})\mathrm{sgn}(l), \hspace{3mm} \text{if} \hspace{3mm} ||\boldsymbol{d}|| > d_{th}
    \\
    0, \hspace{3mm} \text{else}
    \end{cases},
\end{equation}
where $\eta$, $\zeta$, $\beta$, and $k$ $\in$ $\mathbb{R}_{\geq0}$ define the function's maximum and minimum asymptotes, the inflection point, and the steepness of the curve, respectively, and $l$ is the lateral distance between the contact location and the robot center. Finally, the desired linear velocities are computed based on $a_{r}$ and $v^{*}$ as: 
\begin{equation}\label{eq:v_x}
    v_{x}^{R} = v^{*}/{\sqrt{1+a_{r}^2}}, \quad
    v_{y}^{R} = a_{r}v^{*}/\sqrt{1+a_{r}^2}.
\end{equation} 
This formulation enables the robot to produce higher ${v}^{R}_x$ when the contact point is near the robot's center while generating lower ${v}^{R}_y$. Conversely, if the contact approaches the edges of the base, the increase of $a_{r}$ leads to a higher $v_{y}^{R}$, promoting a sliding behavior for the object toward the center of the robot.
The desired angular velocity, $\omega$, that rotates the robot toward the goal, is given as:
\begin{equation}\label{eq:w}
    \omega = v_{x}^{R}\tan({K_{h}(\theta^{*} \ominus \theta)) / L}, \quad
    \theta^{*} = atan2(d_y, d_x),
\end{equation}
where $L$ $\in$ $\mathbb{R}_{>0}$ is a parameter influencing the sharpness of the curvature in the motion, $K_{h}$ $\in$ $\mathbb{R}_{>0}$ is the heading gain, with the heading aligning the robot's X-axis. $\theta$ and $\theta^{*}$ $\in \mathbb{S}$ represent the robot orientation in $\boldsymbol{\Sigma}_{W}$ and the heading angle of $\boldsymbol{d}$, respectively. The $\ominus$ calculates the difference between $\theta$ and $\theta^{*}$, constrained to the range [$-\pi, \pi$).

Despite the dynamic adjustments of lateral movements, if the contact point moves significantly toward the edges, risking a potential loss of contact, the robot activates a Realignment State. 
In this state, $a_{r}$ is used on its maximum to prioritize the ${v}^{R}_y$ over ${v}^{R}_x$, and $\omega$ is reduced to minimize the risk of sliding toward the edges due to the rotational motion. In the case of momentary, though rare, contact loss due to unpredictable object motion or surface irregularities of the ground or object, the robot moves forward at a relatively low velocity ($v_x^R=0.01$ m/s), to immediately re-establish the contact.
A more detailed explanation of these states, along with the pseudocode that demonstrates the flow of the \emph{RPS}, can be found in \cite{idilPushDark}.

\subsection{Preprocessing LiDAR Sensor Data}
\label{leverage_sensor_data_sec}

Efficient preprocessing of 3D LiDAR data is crucial, as raw point clouds are difficult to handle due to their large volume and irregular, sparse structures. In the literature, two commonly used preprocessing approaches are point-based and voxel-based methods. Point-based methods operate each point individually, extracting features such as distance, intensity, and position from the point cloud. However, their high computational cost can result in performance constraints that hinder their use in real-time applications.

In contrast, voxel-based methods partition the 3D space into a fixed grid of cubes (voxels), aggregating the data within each voxel to form a more compact representation.
Many LiDAR-based perception models adopt this voxelization approach, with some even converting 3D point clouds into 2D Bird's-Eye View (BEV) projections to further enhance processing speed and facilitate faster real-time performance \cite{Wang2015VotingFV, PIXOR}. In our approach, both methods described above are adopted, similar to \cite{VoxelNet, FastPointRCNN}, to facilitate the extraction of a compact descriptor with rich feature representations that are well-suited for learning algorithms.

To obtain such a descriptor, the point cloud data in the sensor frame ($\boldsymbol{\mathcal{O}}^{L}\subset\mathbb{R}^3$) is transformed into the robot’s base frame, aligning it with the robot’s coordinate system as $\boldsymbol{\mathcal{O}}^{R} = \boldsymbol{H}^{R}_{L} \boldsymbol{\mathcal{O}}^{L}$, where $\boldsymbol{H}^{R}_{L}$ is the homogeneous transformation of $\boldsymbol{\Sigma}_{L}$ with respect to $\boldsymbol{\Sigma}_{R}$. Then, $\boldsymbol{\mathcal{O}}^{R}$ is filtered to ensure the following steps are limited to only the region of interest, corresponding to the robot’s designated pushing area. The filtered region in front of the robot base is defined as:
\begin{align}
\boldsymbol{\mathcal{O}}^{R,filt} &= \{ \boldsymbol{q}_i^{R} \in \boldsymbol{\mathcal{O}}^{R} \mid \epsilon_x^{min} < q_{i,x}^{R} < \epsilon_x^{max} \land \nonumber \\
& \epsilon_y^{min} < q_{i,y}^{R} < \epsilon_y^{max} \land \epsilon_z^{min} < q_{i,z}^{R} < \epsilon_z^{max} \}, \label{eq:roi}
\end{align}
where \( \boldsymbol{q}_i^{R} \in \mathbb{R}^3 \) are the data points, and \( \epsilon_{*}^{min}, \epsilon_{*}^{max} \in \mathbb{R} \), with \( * \in \{x,y,z\} \), correspond to the scalar boundaries of the filtered region expressed in $\boldsymbol{\Sigma}_{R}$. 
Note that, we set $\epsilon^{min}_y$ and $\epsilon^{max}_y$ as the leftmost and rightmost y-coordinates of the robot base, representing the full pushing region along the Y-axis.
After the filtering process, the 3D point cloud is projected onto a 2D plane ($\boldsymbol{\bar{\mathcal{O}}}^{R,filt}$) using the BEV method by replacing the z-coordinates with a scalar:
\begin{equation}
\boldsymbol{\bar{\mathcal{O}}}^{R,filt} = \{ \bar{\boldsymbol{q}}_i^{R} = [q_{i,x}^{R}, q_{i,y}^{R}, h_{b}]\quad | \quad \boldsymbol{q}_i^{R} \in \mathcal{O}^{R, filt} \}, \label{eq:bev}
\end{equation}
where $h_{b}$ is the height of the robot base.

Following this, the points in $\boldsymbol{\bar{\mathcal{O}}}^{R,filt}$ are grouped to their corresponding \textit{voxel grid cell}, represented by a matrix $\boldsymbol{\mathcal{V}}$ of dimensions $M \times S$, where each entry $\boldsymbol{\mathcal{V}}_{m, s}$ may contain data points by the following condition:
\begin{equation}
\begin{aligned}
    \boldsymbol{\mathcal{V}}_{m,s} &= \{{\bar{\boldsymbol{q}}_i^{R}} \mid a_m \leq\bar{q}_{i, x}^{R}< a_{m+1} \land b_s \leq\bar{q}_{i, y}^{R}< b_{s+1} \}, \\ 
    a_m &= \epsilon_x^{min} +mg_x , \quad \text{for} \quad m = 0, 1, \ldots, M-1,\\
    b_s &= \epsilon_y^{min} +sg_y , \quad \text{for} \quad s = 0, 1, \ldots, S-1,
\end{aligned}
\end{equation}
where $g_x$ and $g_y \in \mathbb{R}_{> 0}$ are the dimensions of the \textit{voxel grid cells}, while $M = (\epsilon^{\text{max}}_x - \epsilon^{\text{min}}_x)/g_x$ and $S = (\epsilon^{\text{max}}_y - \epsilon^{\text{min}}_y)/g_y$ are the total number of rows and columns, respectively. The cell dimensions were carefully tuned based on the following factors: (a) $g_x$ is selected as slightly bigger than the LiDAR beam spacing at the $h_{b}$ plane with a small offset $\delta \ll 1$, ensuring high grid resolution along the X-axis of the robot, and (b) $g_y$ is chosen such that the number of cells along the Y-axis ($S$) matches the number of capacitive-taxels on the robot’s front, as described in \cite{idilPushDark}.

To effectively capture spatial relationships between the object and the robot while reducing the data size for real-time operations, we propose a novel descriptor, proximity vector field ($\{\overrightarrow{\boldsymbol{P}}_s\}_{s=0}^{S-1}$), that builds upon the voxelized point cloud. The starting points of the $\overrightarrow{\boldsymbol{P}_s}$ (see Fig. \ref{fig:framework}) represent the centers of the first cells in each column of $\boldsymbol{\mathcal{V}}$, and it is given by:
\begin{equation}
    \boldsymbol{P}^{start}_s = [\epsilon_x^{min} + g_x/2, \quad \epsilon_y^{min} + g_y(1+2s)/2, \quad h_{b} ].
    \label{eq:P_start}
\end{equation}
\noindent The endpoints of the proximity vectors are then calculated at each control cycle with: 
\begin{equation}
    \boldsymbol{P}^{end}_s = 
    \begin{cases} 
    [ x_s^{min}, P_{s,y}^{start}, P_{s,z}^{start} ], & \text{if } \exists m \text{ s.t. } \boldsymbol{\mathcal{V}}_{m,s} \neq \emptyset \\
    [\epsilon_x^{max}, P_{s,y}^{start}, P_{s,z}^{start} ], & \text{otherwise}
    \end{cases},
    \tag*{\hspace{-5pt}\raisebox{0pt}{(\refstepcounter{equation}\theequation)}}
\end{equation}
where ${x}_s^{min}$ is computed by using the closest occupied \textit{voxel grid cell} to the front side of the robot base, corresponding to $m_{min} = \arg\min \left\{ m \mid \boldsymbol{\mathcal{V}}_{m,s} \neq \emptyset \right\}$ for a given $s$:
\begin{equation}
x_s^{min} = \frac{1}{N} \sum_{i=1}^{N} \bar{q}_{i,x}^{R}, \quad \text{where} \quad \bar{\boldsymbol{q}}_i^{R} \in \boldsymbol{\mathcal{V}}_{m_{min},s}
\end{equation}
where $N$ represents the number of points in $\boldsymbol{\mathcal{V}}_{m_{\text{min}},s}$, \( \bar{q}_{i,x}^{R} \) refers to the x-coordinate of the point \(\bar{\boldsymbol{q}}_i^{R}\) in $\boldsymbol{\Sigma}_{R}$. This formulation allows us to assign the x-component of $\boldsymbol{P}^{end}_{s}$ as the average x-coordinate of the points in the closest cell to the robot for a given $s$. If all the cells in the column $s$ are empty, the x-component of $\boldsymbol{P}^{end}_{s}$ is set to the voxel grid boundary $\epsilon_x^{max}$. Finally, ($\{\overrightarrow{\boldsymbol{P}}_s\}_{s=0}^{S-1}$), as visualized in Fig. \ref{fig:framework}, can be defined as:
\begin{equation}
\overrightarrow{\boldsymbol{P}_{s}} = \boldsymbol{P}^{end}_s - \boldsymbol{P}^{start}_s, \quad \text{for} \quad s = 0, 1, \ldots, S-1.
\end{equation}

The norm of each vector in the proximity field, $\{ ||\overrightarrow{\boldsymbol{P}}_s || \}_{s=0}^{S-1}$, is then used as the input set for the \textit{CPM}, acting as a bridge between the point cloud data and the interpretation of contact between the robot and the pushed object.
The described data preprocessing reduces noise and outliers, yielding a downsampled, smoother, and more consistent representation that enhances its suitability for learning networks.

\subsection{Contact Perception Module (CPM)}\label{CPM_sec}

In this work, a Long Short-Term Memory (LSTM) network, as introduced in \cite{LSTM1997}, is employed to predict the contact state, utilizing its capacity to capture temporal dependencies across consecutive proximity vectors and effectively track changes due to object motion. Moreover, LSTM's gating mechanism helps mitigate the vanishing gradient problem, especially when learning from noisy or sequential data \cite{Pascanu}, making it particularly well-suited for accurate and stable contact estimation.

This module is composed of two sub-networks: the \textit{Contact Location Estimator}, which predicts the point of contact between the object and the pushing surface, and the \textit{Contact Type Estimator}, which classifies the contact as either no contact, point contact or line contact. 
The outputs from these sub-networks are then utilized by the \textit{RPS} to steer the object toward the goal locations.

\subsubsection{Contact Location Estimator (CLE)}
\label{CLE_sec}

The \textit{CLE} is a regression-based LSTM network designed to predict the contact location, using the norm of the proximity vectors ($\{ ||\overrightarrow{\boldsymbol{P}}_s || \}_{s=0}^{S-1}$) as input features that represent the interaction of the robot with an object. It consists of two hidden layers, each containing 32 units, and processes an input sequence of 20 time steps, equivalent to 2 seconds, given that the LiDAR operates at 10 Hz. This sequence length provides the model with sufficient temporal context to capture dependencies in the data effectively.
The fully connected last layer of this LSTM network reduces the output dimension to one, providing the estimated contact location ($\hat{l}$ $\in$ $\mathbb{R}$, as shown in Fig. \ref{fig:framework}) at each control cycle. Both the input and output values are normalized using a min-max scaler \cite{scaler}, which helps to reduce the impact of varying data magnitudes and enables faster convergence and more accurate predictions.

\subsubsection{Contact Type Estimator (CTE)}
\label{CTE_sec}

The \textit{CTE} is a classification LSTM network that takes the same normalized input as the \textit{CLE} ($\{ ||\overrightarrow{\boldsymbol{P}}_s || \}_{s=0}^{S-1}$ with a sequence length of 20), though estimates the contact type ($\hat{\mathcal{C}_{t}}$). It consists of two fully connected hidden layers, each containing 32 hidden units. The final layer generates scalar logits for three possible contact types $\hat{\mathcal{C}_{t}}$ $\in$ \{no contact, point contact, line contact\}. 

\begin{figure*}[!t]	
    \centering
    \includegraphics[width=0.98\textwidth]{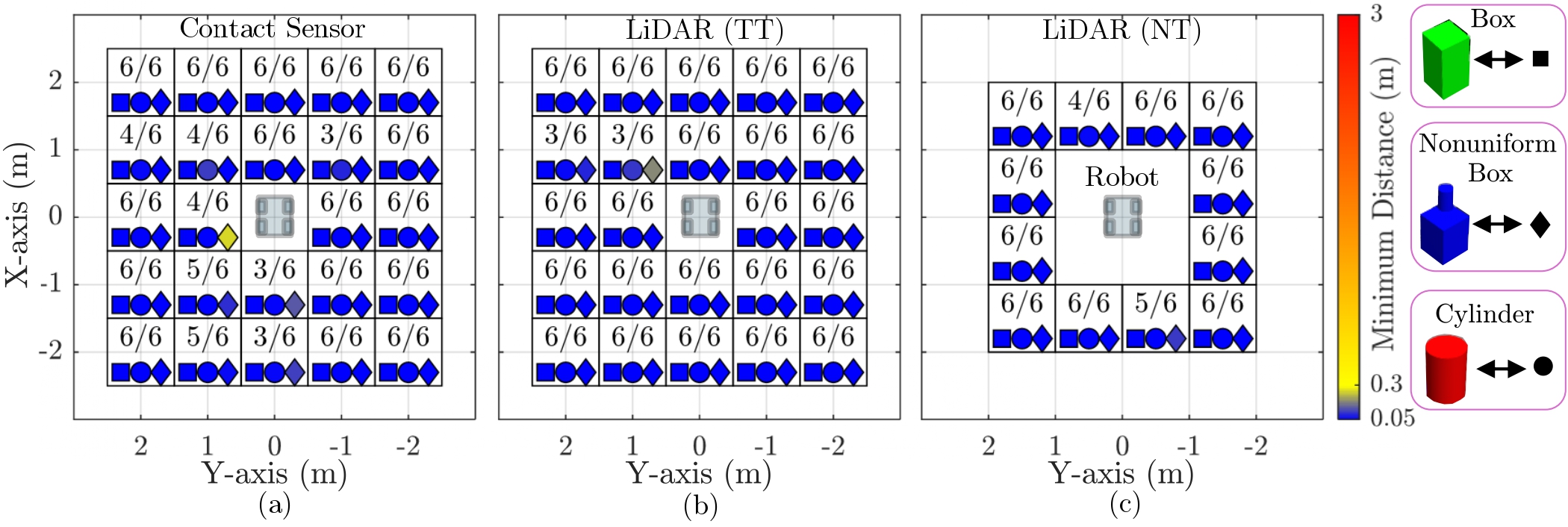}
    \caption{Results of the simulation experiments where \textit{RPS} is pushing objects using (a) a contact sensor, (b) LiDAR toward the same target set (LiDAR TT), and (c) LiDAR toward a new target set (LiDAR NT). The plots show the average minimum distance between the contact point—obtained from the contact sensor in (a) and estimated from LiDAR in (b, c)—and the goal location for both friction sets across all target positions and objects. The pushed objects are: (a) a 20 kg box ($40\times 40\times80$ cm), (b) a 5 kg box ($45\times45\times60$ cm) combined with a 10 kg cylinder (radius 10 cm, height 30 cm) placed at the rear left corner to simulate asymmetric mass distribution, and (c) a 25 kg cylinder (radius 25 cm, height 70 cm).}
    \label{fig:min_distance}
\end{figure*}

\section{Training and Evaluation}

To facilitate the substitution of a specific sensor (e.g., the capacitive touch sensor) with an alternative (e.g., 3D LiDAR sensor) using the proposed approach, we leveraged a simulated environment to generate the data necessary for the training of \textit{CLE} and \textit{CTE} sub-networks. Notably, both sensors being replaced do not need to be physically present, as simulation data alone are sufficient for comprehensive training. This aligns with our motivation of proposing a methodology that allows the given controller to operate alternative sensory data when the original sensor is unavailable or impractical to use.

\subsection{Simulation Environment and System Parameters}\label{subsec:simulation_environment}

The simulation environment was created using Gazebo simulator version 11.11 with the Open Dynamics Engine (ODE). A virtual model of the mobile manipulator was utilized, with the 3D LiDAR mounted on the robot arm's flange. 
The contact sensor provided by Gazebo~\cite{contactsensor} was used to cover the sides of the mobile base, simulating the capacitive touch skin in \cite{idilPushDark}. 
The raw contact sensor outputs were processed to obtain the contact location relative to the robot frame in the X-Y plane ($\boldsymbol{p}^{R}_{C}$ $\in$ $\mathbb{R}^{2}$) and the contact type ($\mathcal{C}_{t}$). 
If the contact positions are confined to an area narrower than 5 cm, the $\mathcal{C}_{t}$ is classified as a point, and $\boldsymbol{p}^{R}_{C}$ is determined as the average position. On the other hand, if the contact region either spans beyond 5 cm or multiple regions are separated by more than this distance, the $\mathcal{C}_{t}$ is categorized as a line. This form of interaction can be considered as if only the extreme points are in contact\cite{Xielinecontact}, and the $\boldsymbol{p}^{R}_{C}$ is computed by averaging the positions of the boundaries. Consequently, the lateral distance between the contact location and the robot center ($l = {p}^{R}_{C, y}$) serves as the target value for the \textit{CLE} network, while the contact type ($\mathcal{C}_{t}$) is employed as the target for the \textit{CTE} network. 

In the experiments, we employed the same set of parameters as in~\cite{idilPushDark} for the \emph{RPS}. 
The span of the region of interest is defined by the following boundaries
$\epsilon_{x}^{min} = 0.3, \, \epsilon_{x}^{max} = 0.62, \, \epsilon_{y}^{min} = -0.3, \, \epsilon_{y}^{max} = 0.3, \, \epsilon_{z}^{min} = -0.4 \, \epsilon_{z}^{max} = 0.5
$  in meters. The \textit{voxel cell} dimensions segmenting this region into a structured grid were set to $g_x = 0.02$ m and $g_y = 0.05$ m.

\subsection{Data Collection and Training}\label{subsec:data_collection_training}

In order to collect a sufficient and diverse dataset for training, objects with varying masses and shapes (see Fig. \ref{fig:min_distance} for their properties) were pushed using contact sensor data as input to the \textit{RPS}, while data from a 3D LiDAR were simultaneously recorded. These experiments were conducted in two distinct virtual environments, each with different sets of friction coefficients: $\boldsymbol{\mathcal{S}}_{\mu1}$ (0.3 between object and ground, 0.35 between object and robot), $\boldsymbol{\mathcal{S}}_{\mu2}$ (0.2 between object and ground, 0.5 between object and robot). For each object-friction pairing, the robot was tasked to push the contact point to 24 target locations on the X-Y plane, spaced 1 meter apart, ranging from -2 to 2 on both axes (excluding the (0,0) start point; see Fig. \ref{fig:min_distance}).
In these trials, the aim is to push the object 0.05 meters (${d}_{succ}$) close to the target within 300 seconds, without losing contact for more than 150 seconds. Since the contact is preserved in the majority of the trials, we also included 12 trials where the robot is initially moved away from the object instead of pushing it to obtain data on scenarios without contact. Altogether, the training set consisted of 156 trials (2 friction coefficients × 24 targets × 3 objects + 12 trials without contact). It is important to note that when pushing toward targets behind the robot, the objects' movement becomes more responsive because of the significant rotation, altering their relative orientation compared to the beginning, especially for the objects with corners. However, this behavior offers valuable data for training the networks to handle such dynamic and less predictable movements.

In the validation dataset, six target points were selected, which are located at coordinates ($\pm$3, $\pm$3) and (0, ±3) meters relative to the robot starting position. These targets were not included in the training phase, allowing us to assess the network’s ability to generalize to unseen goal locations and stop training to prevent overfitting. In total, the validation dataset contained 36 trials (2 friction coefficients $\times$ 6 targets $\times$ 3 objects). Both the \textit{CLE} and \textit{CTE} models were trained on the training dataset with the Adam optimizer \cite{adamOpt}, using a learning rate of 0.001 for 200 epochs. The models with the lowest validation loss, specifically a root-mean-squared error (RMSE) of 0.0136 cm for \textit{CLE} and a cross-entropy loss of 0.04 for \textit{CTE}, were selected for the following experiments. 

\subsection{Simulation Results}
In order to evaluate the effectiveness of the sensor substitution approach during non-prehensile pushing, the previously described simulation environment was utilized. In this setup, the same objects were pushed using the \textit{RPS}, which relied on the $\hat{l}$ and $\hat{\mathcal{C}_{t}}$ derived from the trained \textit{CLE} and \textit{CTE} networks with 3D LiDAR data. These experiments were conducted in two stages: first, the proposed approach was tested on the same target locations as the ones in the training dataset (LiDAR TT), and subsequently, new target locations were used (LiDAR NT).

Fig. \ref{fig:min_distance}a reports the minimum distance achieved between the contact point (where the object touches the robot) and the target location in scenarios where the \textit{RPS} uses the contact sensor data. On the other hand, Fig. \ref{fig:min_distance}b and \ref{fig:min_distance}c demonstrate the same metric based on the $\hat{l}$, when the \textit{RPS} operates using \textit{CPM} estimations. 
In this figure, the minimum distance is shown as the average value calculated across both friction sets at each target position and object. 
In these scenarios, the trial is terminated and considered successful if the minimum distance reaches the ${d}_{succ}$ threshold, which was set to 5 cm (see Sec. \ref{subsec:data_collection_training}). The values for each target position represent the count of successful trials out of a total of 6 trials (consisting of 3 objects and 2 friction sets). In addition to this, the success rates of these experiments for the different object-friction sets are reported in Table \ref{Table:success_Rate}.

\begin{table}[!t]
\caption{Trial Success Rates}
\label{Table:success_Rate}
\centering
\renewcommand{\arraystretch}{0.8} % Reduce row height
\footnotesize
\begin{tabular}{P{1.6cm}C{0.65cm}C{0.65cm}C{0.65cm}C{0.65cm}C{0.65cm}C{0.65cm}}
\toprule 
\multicolumn{1}{c}{} & \multicolumn{2}{c}{Box} & \multicolumn{2}{c}{Cylinder} & \multicolumn{2}{c}{    \begin{tabular}{@{}c@{}} Nonuniform Box\end{tabular}
} \\
\cmidrule(rl){2-3} \cmidrule(rl){4-5} \cmidrule(rl){6-7}
 & {$\mu_{1}$} & {$\mu_{2}$} & {$\mu_{1}$} & {$\mu_{2}$} & {$\mu_{1}$} & {$\mu_{2}$} \\
\midrule
\begin{tabular}{@{}c@{}}Contact Sensor\end{tabular} & $23/24$ & $24/24$ & $19/24$ & $19/24$ & $19/24$ & $23/24$ \\
\midrule
LiDAR (TT) & $24/24$ & $24/24$ & $22/24$ & $22/24$ & $22/24$ & $24/24$  \\
\midrule
LiDAR (NT) & $12/12$ & $12/12$ & $11/12$ & $11/12$ & $11/12$ & $12/12$ \\
\bottomrule
\end{tabular}
\end{table}

The results reveal that the contact sensor can be effectively replaced by the 3D LiDAR using the proposed methodology in the simulation. In fact, the \textit{RPS} strategy utilizing the estimated contact information from the LiDAR shows higher success rates for both the same targets as in the training set (95.83\%) and unseen targets (95.83\%), compared to the one that relies on contact sensor information (88.19\%). This unforeseen increase in the pushing performance when the sensor is replaced can be explained by the small errors between the $\hat{l}$ and the $l$. Although the RMSE error is 0.0415 cm for LiDAR TT and 0.042 cm for LiDAR NT, along with 89.39\% and 90.15\% accuracies in $\hat{\mathcal{C}_{t}}$, respectively, these minor discrepancies in the contact state still influenced the robot's motion.

In order to investigate these results further, we began by analyzing the mean of the absolute values of the $\hat{l}$ and $l$ during the experiments, denoted by $\bar{|l|}$. 
Besides, we defined a metric to represent the normalized angle swept, $\mathrm{\Delta\Theta}_{norm}$, during the trials to understand the extent of rotation executed by the robot while pushing. This metric is calculated as follows:
\begin{equation}
\mathrm{\Delta\Theta}_{norm} = \frac{\sum_{i=0}^{n-1} \left| \theta[t_{i+1}] - \theta[t_i] \right|}{t_{min}-t_{start}},  
\end{equation}
where $\theta$ is the robot orientation in $\boldsymbol{\Sigma}_{W}$, $t_{min}$ is the time the robot reaches the closest point to the target, and $t_{start}$ is the starting time, and $n$ represents the total number of time steps considered in the summation.

\begin{figure}[!t]
\centering
\includegraphics[width=0.98\columnwidth, trim=0cm 0cm 0cm 0cm, clip=true]{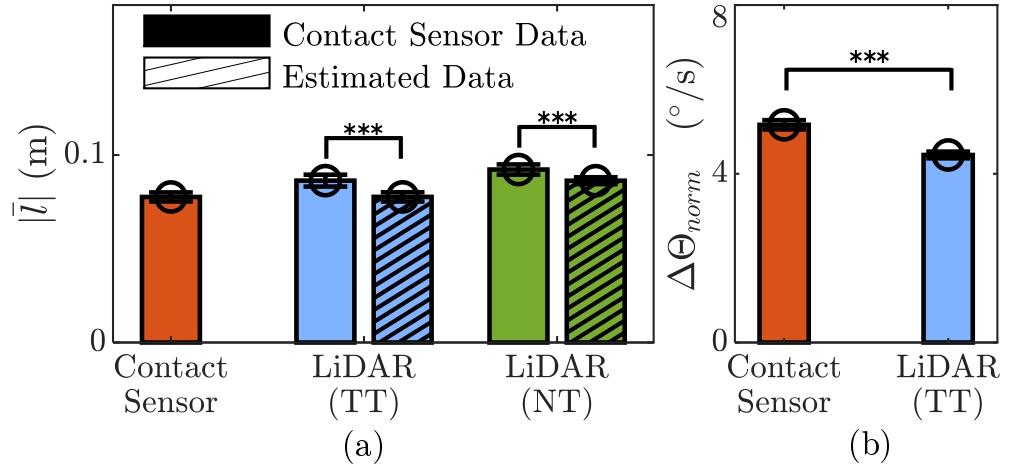}
\caption{(a) The means and standard errors of the absolute values of $l$ and $\hat{l}$, denoted by $|\bar{l}|$,  when \textit{RPS} is pushing the objects with the contact sensor, LiDAR toward the same target set (LiDAR TT), and the new target set (LiDAR NT), and (b) normalized the angle swept when \textit{RPS} relied on contact sensor and the LiDAR for the same target set. The outcomes of the Wilcoxon signed-rank test are represented with: *: $p < 0.05$, **: $p < 0.01$, ***: $p < 0.001$.}
\label{fig:simulation_metrics}
\end{figure}

The results of these metrics are reported in Fig. \ref{fig:simulation_metrics}, with the means and standard errors presented as bar plots, accompanied by the outcomes of the Wilcoxon signed-rank tests. The results are obtained by averaging these measured quantities for all trials. Particularly, Fig. \ref{fig:simulation_metrics}a depicts the $\bar{|l|}$ results for the same set of experiments, where the dashed bar plots show the results for $\hat{l}$ derived from the \textit{CLE} network, and the solid bars correspond to the data obtained from the contact sensor ($l$). In this plot, the $\bar{|l|}$ was not computed for $\hat{l}$ during the contact sensor experiment, as the \textit{RPS} relies solely on the contact sensor data in these scenarios. On the other hand, for the remaining set of experiments, we kept the contact sensor to compare the $\hat{l}$ with the $l$. These results show that the estimated contact locations are statistically significantly closer to the center of the robot than the ones obtained from the contact sensor. The effect of this difference on the robot motion can be seen in Fig. \ref{fig:simulation_metrics}b, which depicts the average $\mathrm{\Delta\Theta}_{norm}$ in the trials. When the \emph{RPS} relied on contact sensor data, the robot exhibited greater rotation while pushing the object toward the target. Conversely, employing the \emph{CPM} reduced the amount of rotation, as the estimated contact locations were closer to the center of the robot compared to the contact sensor data. This reduction in rotational movement can be considered a key factor explaining the observed differences in success rates.
Specifically, it mitigates the "flat bumper problem" \cite{Igarashi2010}, also noted in\cite{idilPushDark}, where excessive rotation causes the pushed object to slide in an undesired direction when a robot with a flat pushing surface was employed.

\section{Framework Scalability}\label{sec:framework_scalability}

Our previous findings demonstrated that replacing a contact sensor with a 3D LiDAR for non-prehensile pushing is not only feasible but highly effective, thanks to the proposed learning-driven approach. Yet, the applicability of this sensor substitution is not confined to the specific experimental setup described so far. The inherent flexibility of the proposed systematic cross-modal transformation approach allows for seamless integration of different sensing modalities, enhancing its adaptability to various robotic platforms and environments.

%In this section, we further validate the proposed approach by employing an RGB-D camera to capture the contact state, notably achieving successful results without the need for retraining the underlying learning models.

In this section, we further validate the proposed approach by employing an RGB-D camera to capture the contact state, notably achieving successful results without the need for retraining the underlying learning models. When LiDAR was used as the substituted sensor, its reliable and accurate 3D point cloud data allowed robust proximity vector extraction. 
In contrast, RGB-D sensors tend to produce noisier and sparser point clouds; however, they offer complementary strengths that can be strategically exploited.
Therefore, instead of relying on the 3D point cloud, we opt to use  depth images in combination with semantic segmentation to derive proximity vectors from RGB-D data.
This extension highlights a key advantage of our approach, which is the ability to adapt to different sensor modalities while preserving performance.  
A prominent aspect of our approach is the use of a descriptor format (the proximity vector), which serves as a modality-agnostic bridge and supports generalization across different sensor types. 
This design not only facilitates the integration of RGB-D cameras but also implies the framework’s potential applicability to configurations without direct depth sensing, such as stereo vision systems, where depth must be inferred via triangulation.

\begin{figure}[!t]	
    \centering
    \includegraphics[width=0.98\columnwidth]{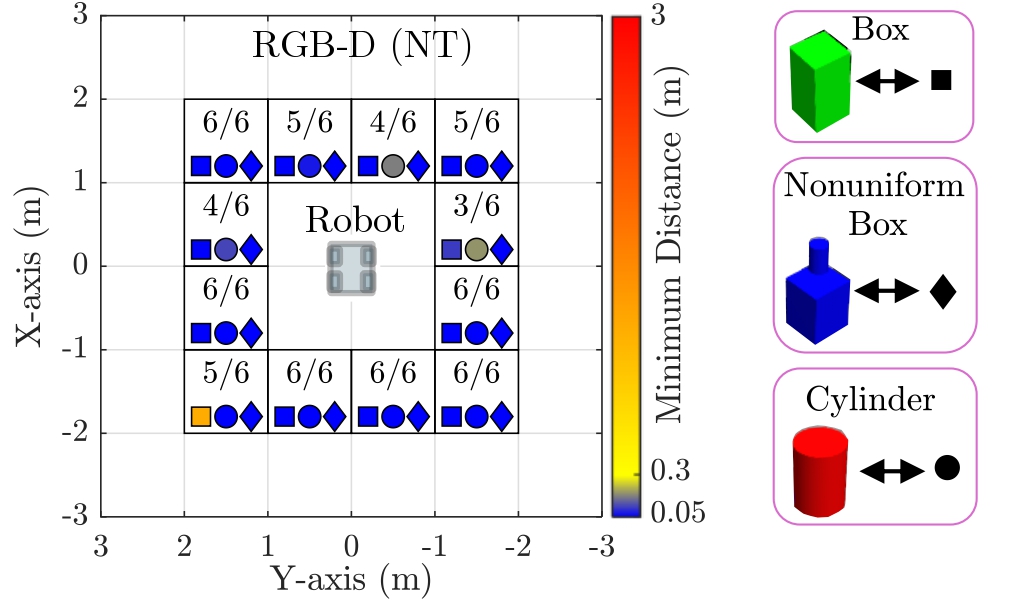}
    \caption{Results of the experiments when RPS is pushing the objects using an RGB-D camera in the simulation environment. The plot shows the average minimum distance between the estimated contact point (from RGB-D) and the goal location, across both friction sets for all target positions and objects.}
    \label{fig:min_distance_rgb}
\end{figure}

\begin{figure*}[!t]	
    \centering
    \includegraphics[width=1\textwidth, trim=0.0cm 0cm 0.0cm 0.0cm, clip=true]{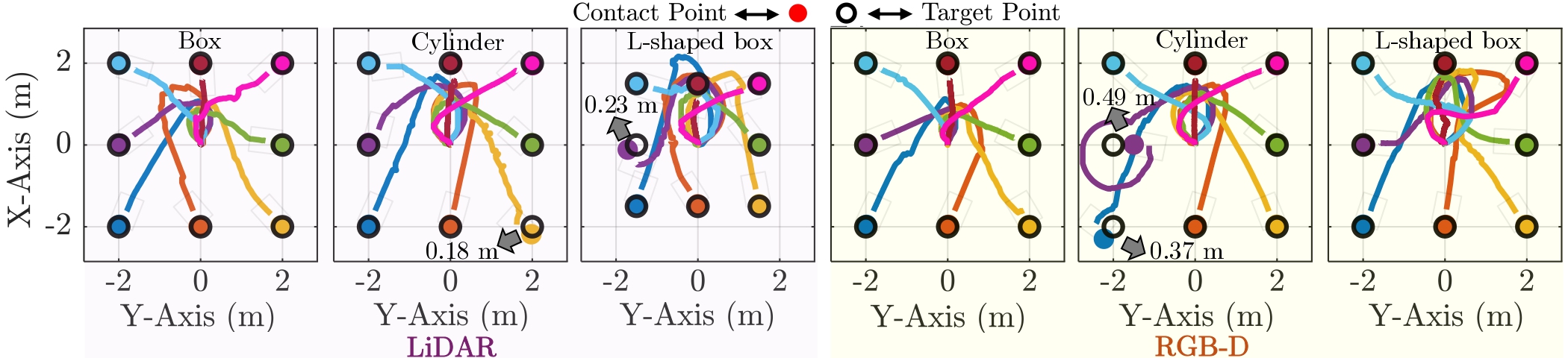}
    \caption{The paths followed by the mobile base, organized by sensory perception modalities (purple and yellow colors) and object types in real-world experiments. The filled circle indicates the object's contact point with the base, while the empty circle denotes the target location.}
	\label{fig:hardware_exps}
\end{figure*}

\subsection{Processing RGB-D Camera Data}\label{subsec:rgb-d processing}

Since the \textit{RPS} strategy generates reactive movements independent of object-specific properties, it is essential to ensure that the method remains robust when handling various objects using an RGB-D camera, just as it did in the case of the 3D LiDAR sensor. This adaptability is crucial for achieving versatile pushing behaviors where objects can vary significantly in shape, size, and texture. 
To this end, we employed the SAM2\cite{sam2} model for object tracking during pushing. With its success in zero-shot segmentation and its scalability across diverse object sizes and shapes, it proves to be an efficient tool for mobile pushing tasks. 
Moreover, its built-in memory mechanism enables continuous tracking of the selected object throughout video streaming. Since the object was positioned in front of the robot at the start of the experiment, we specified the object to be tracked by selecting the pixel value from that area at the beginning. Once selected,  the object remains segmented until the end of the experiment,  even as it slides toward the robot’s edges.

To convert the segmented image frames into the input set of the \emph{CPM} ($\{ ||\overrightarrow{\boldsymbol{P}}_s || \}_{s=0}^{S-1}$) on the fly, we first applied masking to the images to isolate only the segmented region. 
The Canny Edge Detection algorithm \cite{Canny} is then employed to accurately identify the edges of the object, with ($u_i$,$v_i$) representing the pixel coordinates of each edge pixel. These pixels are subsequently transformed into the points in the camera frame $\boldsymbol{r}_i^{C}$ using the intrinsic camera calibration parameters ($f_x, f_y, c_x, c_y$) and the corresponding depth information of the pixels ($z_i$) as follows:
\begin{equation}
    r_{i, x}^{C} = z_i(u_i-c_x)/f_z, \; r_{i, y}^{C} = z_i(v_i-c_y)/f_y,  \; r_{i, z}^{C} = z_i.
\end{equation}

Following this, the set of edge points in the camera frame ($\boldsymbol{\Sigma}_C$), denoted as $\boldsymbol{\mathcal{E}}^{C} \subset \mathbb{R}^3$, is transformed into the $\boldsymbol{\Sigma}_{R}$ as $\boldsymbol{\mathcal{E}}^{R} = \boldsymbol{H}^{R}_{C} \boldsymbol{\mathcal{E}}^{C}$, using the homogeneous transformation matrix $\boldsymbol{H}^{R}_{C}$ that defines $\boldsymbol{\Sigma}_C$ relative to $\boldsymbol{\Sigma}_R$. After the same filtering and projection procedure explained in \eqref{eq:roi} and \eqref{eq:bev}  are performed, we obtain the filtered set of points, denoted by $\boldsymbol{\bar{\mathcal{E}}}^{R,filt}$. To maintain the consistent size of $\{\overrightarrow{\boldsymbol{P}}_s\}_{s=0}^{S-1}$, these points are grouped in different bins ($\boldsymbol{\mathcal{G}}_s$) using the same voxel dimension $g_y$ based on their $r_{i, y}^{R}$ values as follows:
\begin{equation}
\begin{aligned}
    \boldsymbol{\mathcal{G}}_s &= \{\boldsymbol{r}_i^{R}\in \boldsymbol{\bar{\mathcal{E}}}^{R,filt} \mid b_s \leq r_{i, y}^{R}< b_{s+1}\}, \\
    b_s &= \epsilon_y^{min} +sg_y , \quad \text{for} \quad s = 0, 1, \ldots, S-1.
\end{aligned}
\end{equation}

Finally, the endpoints of the proximity vectors (\(\boldsymbol{P}^{end}_s\)) are found at each control cycle as the following rule:
\begin{equation}
    P^{end}_{s,x} =
    \begin{cases} 
        \min \left( \mathrm{r}_{i, x}^{R} \mid \mathbf{r}_i^{R} \in \boldsymbol{\mathcal{G}}_s \right), & \text{if } \boldsymbol{\mathcal{G}}_s \neq \emptyset \\
        \epsilon_x^{max}, & \text{otherwise}
    \end{cases},
\end{equation}
while the $y$ and $z$ components are given by:
\begin{equation}
    P^{end}_{s,y} = P_{s,y}^{start}, \quad P^{end}_{s,z} = P_{s,z}^{start},
\end{equation}
with the same $\boldsymbol{P}^{start}_s$ coordinates as given in \eqref{eq:P_start}.

\subsection{Simulation Results}\label{subsec:simulation_experiments_rgbd}

To validate the adaptability of the sensor replacement approach for non-prehensile manipulation, experiments were performed using an RGB-D without retraining the \textit{CPM}, as discussed previously. In these experiments, the same objects within the described simulation environment were pushed to the target points as in LiDAR NT scenarios (see Fig. \ref{fig:min_distance}c).

Fig. \ref{fig:min_distance_rgb} reports the minimum achieved distance between the $\hat{l}$, which is determined from the RGB-D data, and the target locations.
The calculation of this metric, along with the success rates, follows the same approach as the LiDAR experiments under identical termination conditions of the trials. The results demonstrate the versatility of the proposed sensor substitution approach, which successfully pushed the objects 86.11\% of the targets, including challenging ones behind the robot at the beginning, without retraining. 
In these trials, the proposed approach achieved an RMSE of 0.0879 cm in $\hat{l}$ and predicted whether contact was present or not with 92.79\% accuracy. Although the accuracy in classifying the contact type was  51.95\%, due to the misprediction between point or line contact, the robot's ability to push the object toward targets was not affected as \emph{RPS} does not depend on this distinction.

\section{Real-world Experiments and Results}\label{sec:experiments_and_results}

We assessed the performance of our learning-based strategy through real-world experiments using objects intentionally selected to span a wide range of physical characteristics, such as mass, deformability, friction, and shape: a) a rigid, heavy box (25 kg) with dimensions 32 $\times$ 26 $\times$ 47 cm; b) a rigid, lightweight L-shaped box (3 kg) with longer and shorter sections measuring 80 $\times$ 15 cm and 40 $\times$ 27 cm, respectively, and a height of 45 cm; c) a deformable foam-filled cylinder with a radius of approximately $\approx$ 50 cm, a height of $\approx$ 65 cm, and a total weight of 0.8 kg.
In the experiments, both sensory perception modalities were tested, with one mobile manipulator equipped with a 3D LiDAR and the other employing an RGB-D camera to evaluate our approach in a sim-to-real setup.
The video of the real-world experiments is available in the multimedia extension and at \url{https://youtu.be/6yIRcfn2DsY}, showcasing an additional scenario where a box containing a 0.9-kg ball dynamically changes its mass distribution as it rolls inside, while the box is externally disturbed as well during the task.

Fig. \ref{fig:hardware_exps} illustrates the paths of the mobile bases, goal positions, and estimated contact points at the end of the task for all sensory perception modalities and the objects.
The box and the cylinder were pushed using the mobile manipulator equipped with LiDAR, to the target points defined by the coordinates (±2, ±2) and (0, ±2), while the L-shaped object was pushed to closer ones (±1.5, ±1.5) and (0, ±1.5) to evaluate whether the first target set represented the closest feasible limit for pushing in the robot's vicinity.
It was observed that the performance remained consistent across both target sets, achieving an overall success rate of 91.67\%. Furthermore, the success rate of the scenarios remained the same when the RGB-D camera was used instead of LiDAR, confirming the system's scalability to integrate new sensors without retraining.
Moreover, analysis of the unsuccessful trials showed the mean contact-to-target distance was 0.31$\pm$0.12 m, missing success by a small margin of about 25 cm.

\section{Discussion and Conclusion}
\label{sec:discussion_and_conclusion}

In this paper, we proposed a sensor substitution framework for adaptive nonprehensile mobile manipulation. Our approach enables cross-modal transformation between missing and substitute modalities, using a compact descriptor and LSTM networks trained solely on simulation data. The \textit{Contact Perception Module} successfully mimicked contact sensor outputs using LiDAR, achieving over 95\% pushing success rates. Even when replacing LiDAR with an RGB-D camera, performance dropped by only 10\%, demonstrating adaptability without retraining.
Real-world tests across diverse robot-sensor-object setups confirmed over 90\% success rates, even for challenging targets. The system consistently performed well across hardware variations, meeting the real-time requirements of the task without the need for additional tuning. Substituting the contact sensor in \cite{idilPushDark} expanded the range of manipulable objects, improving versatility.
This work demonstrated sensor substitution in adaptive mobile manipulation, reducing hardware dependency while maintaining robust performance, paving the way for more flexible and cost-effective robotic systems.

Future work will focus on enhancing the robustness and generalizability of the proposed framework by assessing its performance on objects with more complex physical characteristics. In particular, we intend to address challenges associated with objects that have transparent, semi-transparent, or highly reflective surfaces, which pose difficulties for RGB-D cameras and LiDAR. In these scenarios, ultrasonic sensors could provide a viable alternative, allowing for the manipulation of such objects, as they are inherently unaffected by surface characteristics. Accordingly, we plan to extend the current cross-modal sensor substitution approach to incorporate additional sensing modalities.

\bibliographystyle{IEEEtran}
\bibliography{biblio}

\begin{IEEEbiography}[{\includegraphics[width=1in,height=1.25in,clip,keepaspectratio]{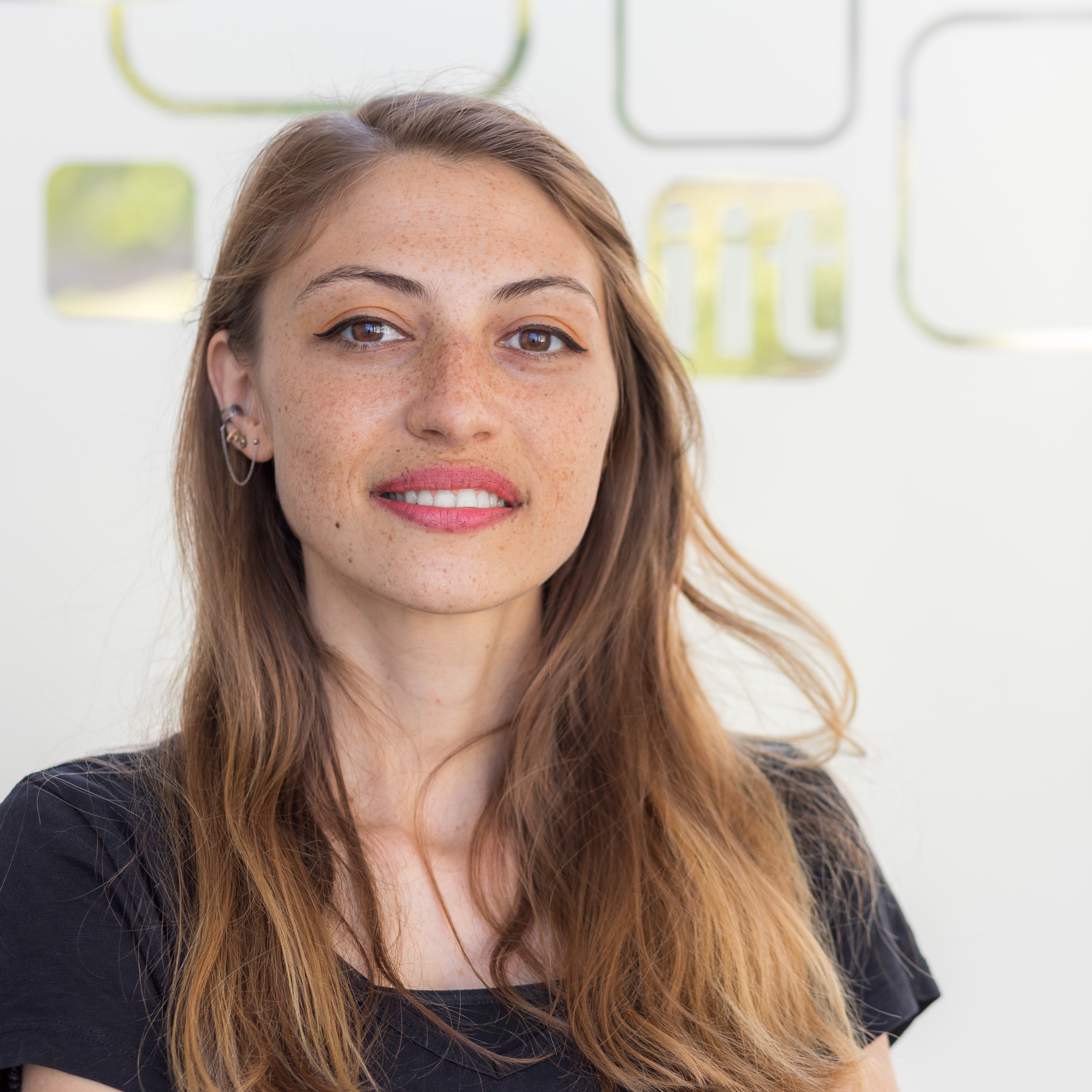}}]{Idil Ozdamar} received the B.Sc. in electrical and electronics engineering and the M.Sc. in biomedical sciences and engineering from Koç University, Istanbul, in 2018 and 2020, respectively. She is currently pursuing the Ph.D. degree with the Human-Robot Interfaces and Interaction Lab (HRI$^2$), Istituto Italiano di Tecnologia (IIT), and at the Department of Informatics, Bioengineering, Robotics, and System Engineering, University of Genoa, Italy. 

Her main research interests include human-robot interaction, haptic perception, collaborative robots, human-in-the-loop control, and contact-rich manipulation.
\end{IEEEbiography}

\begin{IEEEbiography}[{\includegraphics[width=1in,height=1.25in,clip,keepaspectratio]{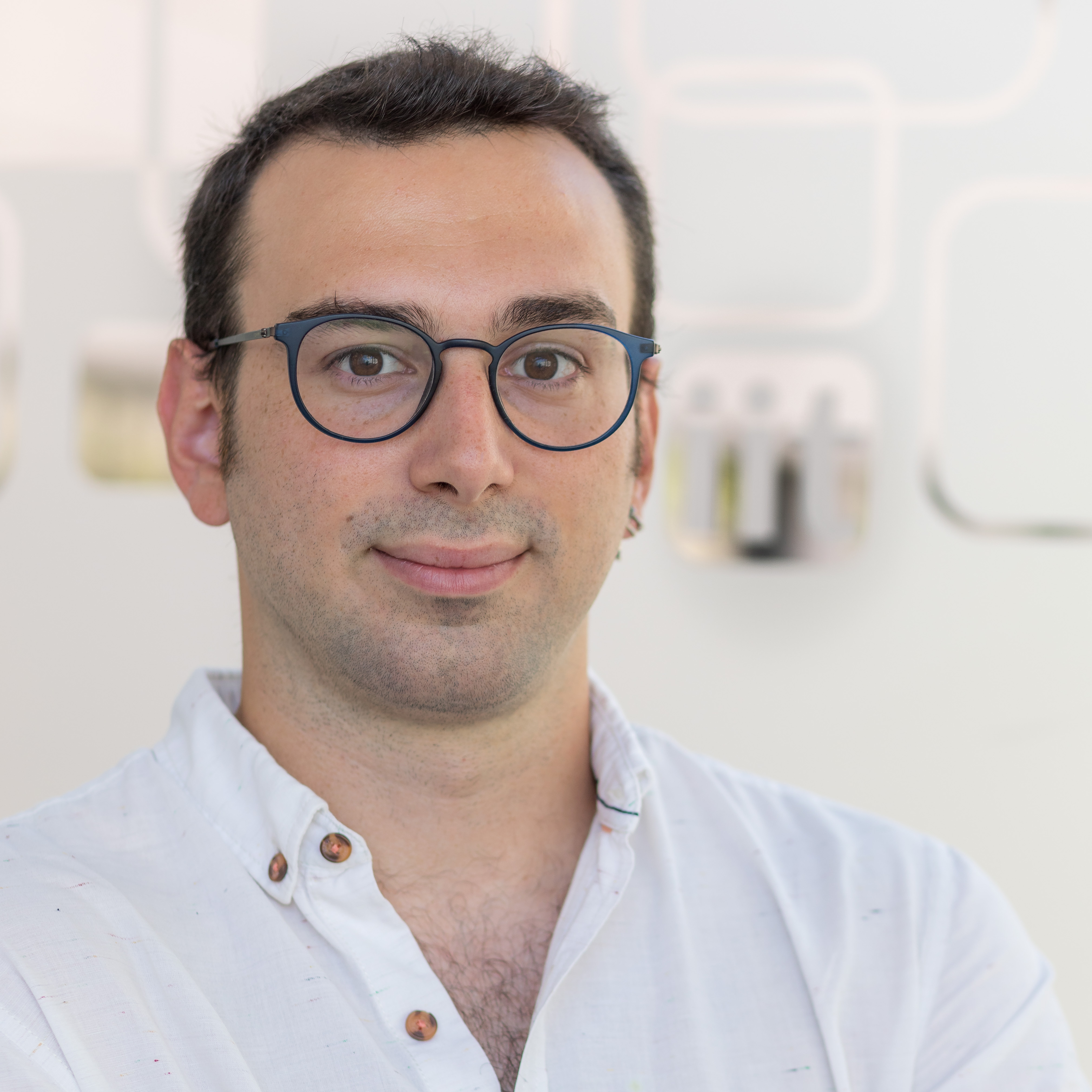}}]{Doganay Sirintuna} received the B.Sc. and M.Sc. degrees in mechanical engineering from Koç University, Istanbul, in 2018 and 2020, respectively. He is currently pursuing the Ph.D. degree with the Human-Robot Interfaces and Interaction Lab (HRI$^2$), Istituto Italiano di Tecnologia (IIT), and at the Department of Informatics, Bioengineering, Robotics, and System Engineering, University of Genoa, Italy. 

His research interests include physical human-robot interaction, mobile manipulation, and robot learning.

Mr. Sirintuna was granted the Academic Excellence Award at his graduation from Koç University in 2020. 

\end{IEEEbiography}

\begin{IEEEbiography}[{\includegraphics[width=1in,height=1.25in,clip,keepaspectratio]{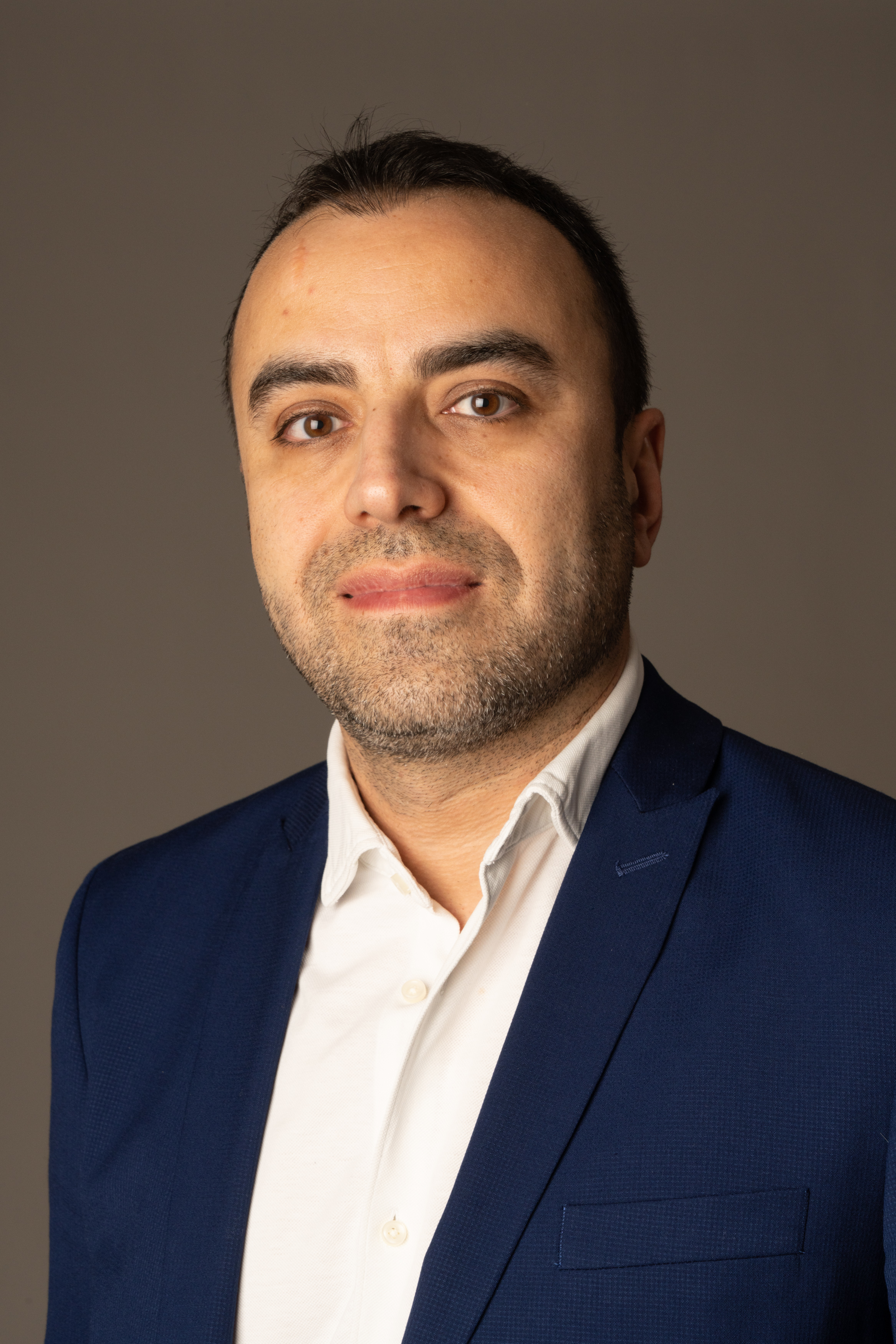}}]{Arash Ajoudani} received the Ph.D.\ degree in robotics and automation from the University of Pisa and the Italian Institute of Technology (IIT), Italy, in 2014. 

He is currently the Director of the Human-Robot Interfaces and Interaction (HRI$^2$) Laboratory at IIT. He holds two ERC grants (Real-Move, 2023; Ergo-Lean, 2019) and serves as the Coordinator or Co-Coordinator of several European projects, including H2020 SOPHIA and CONCERT. He is also a Principal Investigator in various other European and national projects, such as Tornado, RAICAM, LABORIUS, COROMAN, and ReFinger. He is the author of the book \emph{Transferring Human Impedance Regulation Skills to Robots} (Springer, 2020) and has published extensively on physical human-robot interaction, mobile manipulation, robust control, assistive robotics, and tele-robotics.

Dr.\ Ajoudani is a recipient of the IEEE RAS Early Career Award (2021) and the MECSPE Robotics and AI Award (2025). He was a finalist for the Georges Giralt Ph.D.\ Award (2015) and has received multiple best paper award nominations at ICRA, IROS, and Humanoids. He is currently a Senior Editor of the \emph{International Journal of Robotics Research} (IJRR).
\end{IEEEbiography}

\end{document}